\documentclass[letterpaper, 10 pt, conference]{ieeeconf}  

\usepackage{graphicx} 
\usepackage{float}    
\usepackage{caption}  
\usepackage{amsmath}
\usepackage{url}
\usepackage{amssymb}

\IEEEoverridecommandlockouts                              

\title{\LARGE \bf
Retrieval-Augmented Multimodal Depression Detection}

\author{
    Ruibo Hou$^{1,\dagger}$, Shiyu Teng$^{1,\dagger}$, Jiaqing Liu$^{1}$, Shurong Chai$^{1}$, Yinhao Li$^{1}$, Lanfen Lin$^{2}$ \\ 
    and *Yen-Wei Chen$^{1}$, \emph{Member, IEEE}
    \thanks{$^{1}$Ruibo Hou, Shiyu Teng, Jiaqing Liu, Shurong Chai, Yinhao Li and Yen-Wei Chen are with the College of Information Science and Engineering, Ritsumeikan University, Osaka, Japan.}
    \thanks{$^{2}$Lanfen Lin is with the College of Computer Science and Technology, Zhejiang University, Hangzhou, China.}
    \thanks{$^{\dagger}$These authors contributed equally to this work.}
    \thanks{*Corresponding Author: Yen-Wei Chen (chen@is.ritsumei.ac.jp)}
}

\begin{document}

\maketitle
\thispagestyle{empty}
\pagestyle{empty}

\begin{abstract}

Multimodal deep learning has shown promise in depression detection by integrating text, audio, and video signals. Recent work leverages sentiment analysis to enhance emotional understanding, yet suffers from high computational cost, domain mismatch, and static knowledge limitations. To address these issues, we propose a novel Retrieval-Augmented Generation (RAG) framework. Given a depression-related text, our method retrieves semantically relevant emotional content from a sentiment dataset, and uses a Large Language Model (LLM) to generate an Emotion Prompt as an auxiliary modality. This prompt enriches emotional representation and improves interpretability. Experiments on the AVEC 2019 dataset show our approach achieves state-of-the-art performance with CCC of 0.593 and MAE of 3.95, surpassing previous transfer learning and multi-task learning baselines.

\end{abstract}

\section{INTRODUCTION}

The rising prevalence of depression has become a major global health concern, significantly impacting individuals and straining healthcare systems~\cite{friedrich2017depression}~\cite{santomauro2021global}. With limited time and resources available for psychologists to treat each patient, there is an urgent need for automated depression detection using deep learning to improve diagnostic efficiency and accuracy. 

\begin{figure}[t] 
    \centering
    \includegraphics[width=\linewidth]{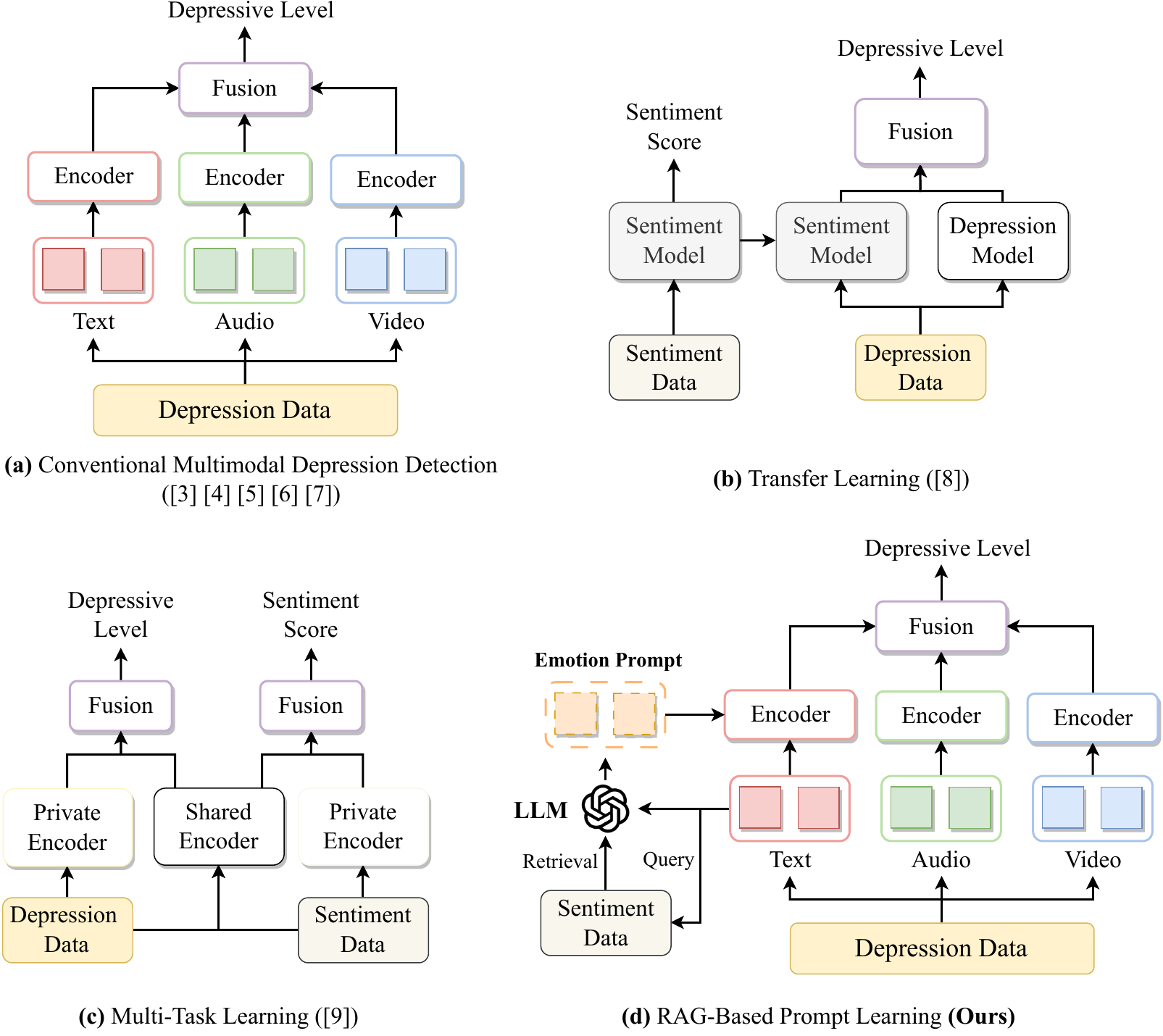} 
    \caption{Different multimodal depression detection frameworks. Unlike previous methods, our approach introduces the Emotion Prompt as a novel modality. By dynamically retrieving semantically similar text and sentiment labels from an external emotional dataset and integrating them with the original text, the Emotion Prompt serves as a fourth modality alongside text, audio, and video. This enables a more balanced multimodal fusion, significantly enhancing the representation of emotional cues and overall model performance.}
    \label{fig:overview}
\end{figure}

In depression detection using deep learning, multimodal data fusion, which integrates video, audio, and text, has gained considerable attention for its accuracy and robustness. Fig.~\ref{fig:overview} outlines various frameworks for multimodal depression detection. 
As Fig.~\ref{fig:overview}(a) illustrates, compared to single-modality approaches, combining linguistic content, vocal features, and facial expressions offers a more comprehensive understanding of an individual's mental health state~\cite{yin2019multi}~\cite{sun2021multi}~\cite{liu2022computer}~\cite{teng2022transformer}~\cite{sun2022tensorformer}. However, training end-to-end networks directly on small-scale depression datasets often faces challenges such as overfitting and insufficient representation learning from weaker modalities.

To address these limitations, researchers have explored the relationship between sentiment analysis tasks and depression detection tasks, leveraging large-scale sentiment datasets to learn generalized emotional representations before transferring them to depression-specific tasks~\cite{10782904}~\cite{10444213}. As presented in Fig.~\ref{fig:overview}(b) and (c), two prominent approaches have emerged in this context. The pre-training and fine-tuning strategy~\cite{10782904} involves first training models on large-scale sentiment datasets, such as CMU-MOSEI~\cite{zadeh2018multimodal}, to acquire generalized multimodal emotional representations, followed by fine-tuning on depression-specific datasets to extract task-specific features for accurate detection. In contrast, the multi-task learning approach~\cite{10444213} designs frameworks that simultaneously learn sentiment analysis and depression detection tasks, enabling sentiment-related features to assist in depression prediction and thereby improving model generalization. These strategies, based on sentiment dataset transfer learning and multi-task learning, have shown measurable improvements in both the accuracy and stability of depression detection models. However, these approaches still face several limitations. First, the training process is often computationally intensive and time-consuming, requiring extensive pre-training or joint training on large-scale sentiment datasets. Second, there is a notable domain discrepancy between sentiment datasets and depression datasets, as emotional expressions in general sentiment analysis may not fully align with those observed in clinical depression scenarios. Third, the knowledge acquired during pre-training tends to be static, limiting the model's ability to dynamically adapt to new inputs or context-specific emotional cues.

\begin{figure*}[t] 
    \centering
    \includegraphics[width=1.0\linewidth]{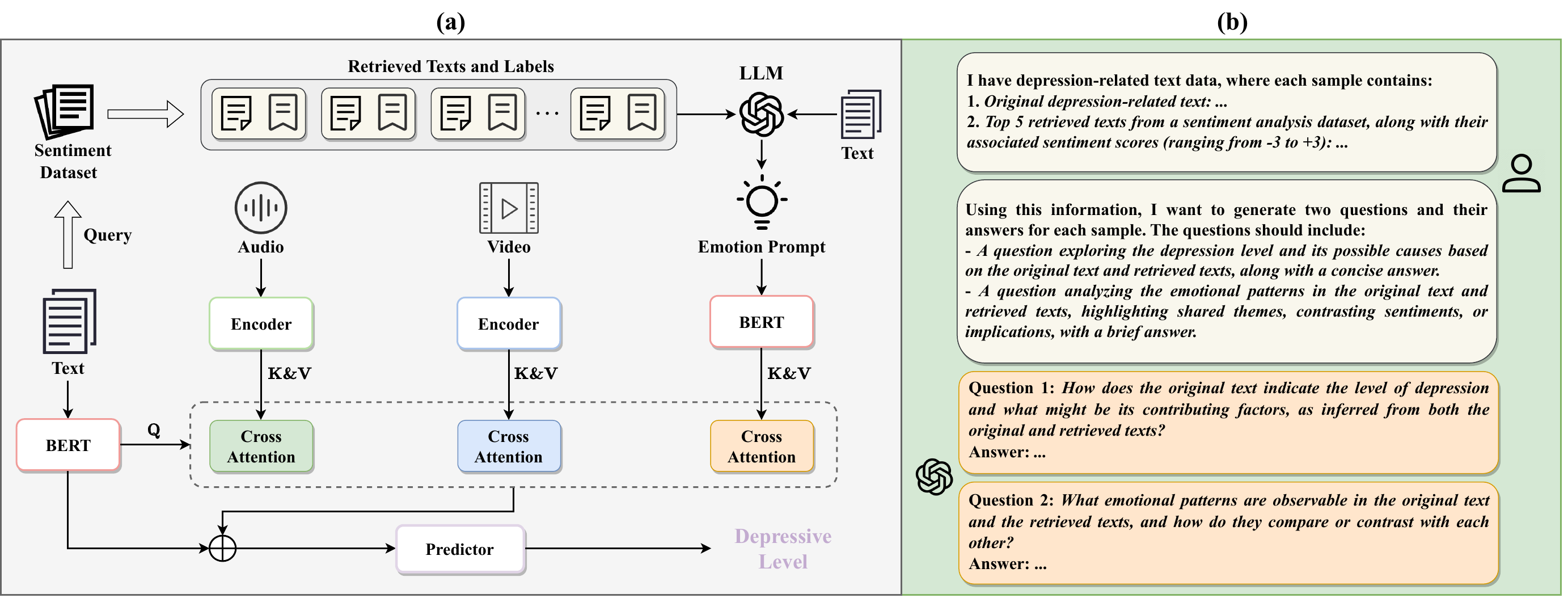} 
    \caption{(a) An overview of our proposed framework, where depression-related text and the Emotion Prompt are fed into the same pre-trained BERT model, while the audio and video encoders are trained from scratch. (b) The process of interacting with the LLM to generate the Emotion Prompt.}
    \label{fig:framework}
\end{figure*}

To further address the aforementioned limitations, this study introduces a Retrieval-Augmented Generation (RAG)-based framework (Fig.~\ref{fig:overview}(d)) for multimodal depression detection, designed around dynamic retrieval of external emotional data and Emotion Prompt generation. RAG methods~\cite{lewis2020retrieval}~\cite{izacard-grave-2021-leveraging} enhance Large Language Models (LLMs) by integrating retrieved knowledge to improve accuracy and mitigate hallucinations. Inspired by this, unlike traditional pre-training and fine-tuning or multi-task learning approaches, our method eliminates the need for extensive pre-training on large-scale datasets. Instead, during analysis of depression-related text, the model dynamically retrieves semantically relevant emotional content and corresponding sentiment labels from an external emotional dataset. These retrieved elements, combined with the original depression text, are fed into GPT-4~\cite{openai2023gpt4} to generate an Emotion Prompt. This prompt serves as a fourth modality, alongside text, audio, and video, enabling a more balanced multimodal fusion and enhancing the representation of emotional cues from weaker modalities.

Compared to conventional transfer learning or multi-task learning frameworks, this retrieval-augmented generation strategy dynamically integrates the most relevant emotional knowledge during runtime. This approach avoids reliance on static pre-trained weights, which often struggle to adapt to task-specific emotional contexts. While we also rely on emotional datasets as external resources, the retrieval mechanism ensures that the model selectively retrieves contextually relevant samples on demand, rather than indiscriminately leveraging global pre-training. This reduces training complexity, mitigates cross-domain distribution mismatches, and enhances the adaptability of the model to dynamic emotional cues.

The main contributions of this study are as follows:  
\begin{enumerate}
    \item Introduction of an RAG + Prompt multimodal fusion framework:  
    Depression-related text serves as a query to dynamically retrieve semantically similar emotional text and sentiment labels from an external sentiment dataset. These are combined with the original text and fed into an LLM to generate an Emotion Prompt. This prompt, acting as a fourth modality alongside text, audio, and video, facilitates cross-attention interactions, enhancing the representation of emotional cues across modalities.

    \item Overcoming the limitations of pre-training:  
    Through dynamic retrieval and generation, our framework minimizes reliance on extensive pre-training while effectively reducing cross-domain distribution mismatches. Unlike static pre-trained weights, dynamic retrieval allows the model to adaptively select contextually relevant emotional cues, ensuring better alignment with depression-related data.

    \item Improved detection performance and interpretability:  
    Experimental results on the AVEC 2019~\cite{ringeval2019avec} benchmark dataset demonstrate that our method achieves state-of-the-art accuracy. Moreover, the explicit retrieval process and the structured Emotion Prompt provide enhanced interpretability, offering valuable insights for clinical diagnosis and decision-making.
\end{enumerate}

\section{METHODOLOGY}


The overview of the proposed method is shown in Fig.~\ref{fig:framework}(a). We employ a RAG approach, where depression-related text serves as a query to retrieve semantically similar text and corresponding sentiment scores from an external sentiment dataset. The retrieved results are combined with the original depression text and fed into an LLM to dynamically generate an Emotion Prompt. Serving as an additional modality alongside text, audio, and video, the Emotion Prompt contributes to the final depression level prediction by enhancing emotional cue representation and enabling more balanced multimodal fusion. Details of this overall framework are described in \textsection\ref{subsection:overview}.

The specific interaction with the LLM to generate the Emotion Prompt is illustrated in Fig.~\ref{fig:framework}(b). During this process, retrieved emotional text and sentiment labels are contextualized and processed by the LLM to construct a structured Emotion Prompt. This step bridges the semantic gap between depression-related and external emotional data, further enriching the model's emotional understanding. Further details are provided in \textsection\ref{subsection:rag_prompt}.

\subsection{The Proposed Framework: An Overview}
\label{subsection:overview}

Firstly, in a depression dataset \( D \), each sample contains three modalities: text (\( t \)), audio (\( a \)), and video (\( v \)). The text modality \( t \) is processed through a BERT~\cite{devlin-etal-2019-bert} encoder pre-trained on extensive textual data, while the audio \( a \) and video \( v \) modalities are passed through randomly initialized encoders. This produces corresponding embeddings: \( h_t \), \( h_a \), and \( h_v \).

Using an external sentiment dataset \( S \), the text embedding \( h_t \) serves as a query to retrieve the top-\( K \) most similar texts, denoted as \(\{s_i\}_{i=1}^{K}\),
based on embedding similarity, accelerated by FAISS~\cite{johnson2019billion} in an offline retrieval process. These retrieved texts, along with their corresponding sentiment scores \( \{y_{s_i}\}_{i=1}^{K} \), are combined with the original text \( t \) and fed into a GPT-4 model to generate an Emotion Prompt \( p \). This prompt is then passed through the same pre-trained BERT encoder to obtain its embedding \( h_p \).

In the subsequent cross-attention mechanism, the text embedding \( h_t \) serves as the Query ($\mathbf{Q}$), while each modality embedding—audio \( h_a \), video \( h_v \), and the Emotion Prompt \( h_p \)—is independently used as Keys ($\mathbf{K}$) and Values ($\mathbf{V}$) to produce modality-specific outputs:

\begin{equation}
\begin{aligned}
    h_{\text{a}} &= \text{CrossAttention}(h_t, h_a, h_a), \\
    h_{\text{v}} &= \text{CrossAttention}(h_t, h_v, h_v), \\
    h_{\text{p}} &= \text{CrossAttention}(h_t, h_p, h_p).
\end{aligned}
\end{equation}

These modality-specific outputs are then concatenated with the original text embedding \( h_t \) to form the final multimodal representation:

\begin{equation}
h_{\text{final}} = \text{Concat}(h_t, h_{\text{a}}, h_{\text{v}}, h_{\text{p}}).
\end{equation}

This representation is passed into a predictor for depression level prediction. The loss function minimizes statistical discrepancies:

\begin{equation}
L_{\text{dep}} = 1 - \frac{2 s_{y\hat{y}}}{s_{\hat{y}}^2 + s_{y}^2 + (\bar{\hat{y}} - \bar{y})^2},
\end{equation}
where \( \bar{\hat{y}} \) and \( \bar{y} \) represent the mean values of the predicted and ground truth distributions, respectively. \( s_{\hat{y}}^2 \) and \( s_{y}^2 \) denote the variances of the predicted and ground truth distributions, while \( s_{y\hat{y}} \) indicates the covariance between them. This loss function minimizes the discrepancy between predicted and ground truth distributions by aligning their mean, variance, and covariance, offering a statistically robust optimization objective.

\subsection{Emotion Prompt via Retrieval-Augmented Generation}
\label{subsection:rag_prompt}

As shown in Fig.~\ref{fig:framework}(b), during the process of guiding the LLM to generate the Emotion Prompt, we first provide the original depression-related text along with semantically similar sentences and their sentiment labels retrieved from an external sentiment dataset. Based on these inputs, the LLM is instructed to:

- Formulate a question to assess the depression level and its potential causes by analyzing both the original text and the retrieved emotional content. This guides the model to uncover explicit and implicit depressive cues and better understand the contributing factors.

- Construct another question to analyze emotional patterns across the original and retrieved texts, focusing on shared themes, contrasting sentiments, and underlying emotional implications. This cross-text emotional analysis bridges semantic gaps, aligns retrieved emotional cues with the original context, and enriches the representation of emotional dependencies.

This design enhances the extraction of emotional cues and reduces ambiguity in emotional interpretation. By generating structured questions and answers, the LLM facilitates a deeper semantic understanding, captures subtle emotional patterns, and bridges contextual dependencies. This improves the overall quality of the Emotion Prompt, contributing to more accurate and interpretable depression severity predictions.

\section{EXPERIMENTAL SETTINGS}
\subsection{Dataset}

\subsubsection{AVEC 2019 Dataset}

The AVEC 2019 Depression Detection Sub-Challenge (DDS) Dataset~\cite{ringeval2019avec} consists of audiovisual recordings of clinical interviews conducted with a virtual agent, designed to assess depression levels using PHQ-8 scores. The accompanying audio recordings have been transcribed using Google Cloud's speech recognition service and annotated with both verbal and nonverbal features. Previous studies, such as those by Sun \textit{et al.}~\cite{sun2021multi}, have identified Mel-Frequency Cepstral Coefficients (MFCC) as the most discriminative acoustic feature and Action Unit (AU) poses as the most informative visual feature. To maintain computational efficiency and simplify the feature extraction process, we focus exclusively on MFCC and AU-poses in our depression detection framework. The dataset includes detailed annotations, such as PHQ-8 scores, interview identifiers, binary depression labels, and gender information. It comprises 163 samples for training, 56 for validation, and 56 for testing.

\subsubsection{CMU-MOSEI Dataset}

The CMU-MOSEI dataset~\cite{zadeh2018multimodal} is a large-scale multimodal dataset curated for sentiment and emotion analysis. It contains 23,454 video clips sourced from YouTube movie reviews, with sentiment scores ranging from -3 (strongly negative) to 3 (strongly positive), annotated by human evaluators. The dataset is divided into 16,315 utterances for training, 1,817 for validation, and 4,654 for testing. In this study, we utilize only the \textit{textual modality} of the CMU-MOSEI dataset as an external resource for retrieval-based augmentation. This textual data serves as the foundation for generating the Emotion Prompt, enabling a dynamic integration of sentiment cues into the depression detection process.

\subsection{Implementation Details}

The model is optimized using the Adam optimizer with a batch size of 32. Training is conducted for 400 epochs, starting with an initial learning rate of \( 6 \times 10^{-4} \), which is reduced by a factor of 0.1 every 100 epochs. Additionally, the learning rate for the BERT encoder is set to one-tenth of the learning rate applied to the rest of the model components. For text encoding, we utilize the pre-trained \textit{bert-base-uncased} model\footnote{\url{https://huggingface.co/google-bert/bert-base-uncased}}. The number of retrieved texts, denoted as \( K \), is set to 5. 

\subsection{Evaluation Metrics}

In depression detection tasks, the \textit{concordance correlation coefficient (CCC)}~\cite{lawrence1989concordance} is widely adopted to quantify the consistency between predicted depression severity scores and the actual ground truth scores in regression scenarios. The CCC is mathematically expressed as:

\begin{equation}
\text{CCC} = \frac{2 s_{y\hat{y}}}{s_{\hat{y}}^2 + s_{y}^2 + (\bar{\hat{y}} - \bar{y})^2},
\end{equation}
where \( \bar{\hat{y}} \) and \( \bar{y} \) represent the mean values of the predicted and ground truth scores, respectively. The terms \( s_{\hat{y}}^2 \) and \( s_{y}^2 \) correspond to the variances of the predicted and actual values, while \( s_{y\hat{y}} \) denotes their covariance. CCC values range from \(-1\) to \(1\), where a value of \(-1\) indicates a complete negative correlation, and \(1\) signifies perfect agreement between predictions and ground truth scores.

In addition to CCC, we also employ the \textit{mean absolute error (MAE)} to evaluate model performance, defined as:

\begin{equation}
\text{MAE} = \frac{1}{N} \sum_{i=1}^{N} \left| \hat{y}_i - y_i \right|,
\end{equation}
where \( N \) denotes the total number of samples, \( \hat{y}_i \) represents the predicted value for the \( i \)-th sample, and \( y_i \) is its corresponding ground truth value. The MAE captures the average magnitude of errors by calculating the absolute difference between predicted and true values across all samples.

These two metrics complement each other, with CCC focusing on the alignment of predicted and actual distributions, while MAE provides a straightforward assessment of prediction accuracy by averaging the deviation across individual samples.

\section{EXPERIMENTAL RESULTS}
\subsection{Performance Comparisons}

As shown in Table~\ref{tab:sota_comparison}, our proposed method achieves the best performance across both evaluation metrics, with a CCC of \textbf{0.593} and an MAE of \textbf{3.95}. These results significantly surpass all other state-of-the-art methods, highlighting the superiority of our approach in depression severity prediction.

\begin{table}[h!]
\centering
\caption{Comparison with state-of-the-art methods.}
\label{tab:sota_comparison}
\begin{tabular}{lcc}
\hline
\textbf{Method} & \textbf{CCC ↑} & \textbf{MAE ↓} \\ 
\hline
AVEC 2019~\cite{ringeval2019avec} & 0.111 & 6.37 \\
EF~\cite{kaya2019predicting} & 0.344 & - \\
Bert-CNN \& Gated-CNN~\cite{rodrigues2019multimodal} & 0.403 & 6.11 \\
Hierarchical BiLSTM~\cite{yin2019multi} & 0.442 & 5.50 \\
Adaptive Fusion Transformer~\cite{sun2021multi} & 0.443 & 5.66 \\
Teng \textit{et al.} (Multi-Task Learning) ~\cite{10444213} & 0.466 & 5.21 \\
Tensorformer~\cite{sun2022tensorformer} & 0.493 & 5.13 \\
Teng \textit{et al.} (Transfer Learning) ~\cite{10782904} & 0.507 & 4.77 \\
\hline
\textbf{Ours} & \textbf{0.593} & \textbf{3.95} \\
\hline
\end{tabular}
\end{table}

Compared to methods trained solely on depression datasets, such as AVEC 2019~\cite{ringeval2019avec}, EF~\cite{kaya2019predicting}, Bert-CNN \& Gated-CNN~\cite{rodrigues2019multimodal}, Hierarchical BiLSTM~\cite{yin2019multi}, Adaptive Fusion Transformer~\cite{sun2021multi}, and Tensorformer~\cite{sun2022tensorformer}, our method demonstrates significant improvements in both CCC and MAE. This demonstrates the effectiveness of our multimodal fusion strategy and the integration of external emotional cues.

Furthermore, when compared with methods leveraging sentiment datasets, such as Teng \textit{et al.} (Multi-Task Learning)~\cite{10444213} and Teng \textit{et al.} (Transfer Learning)~\cite{10782904}, our approach achieves notable performance gains. Unlike traditional methods that rely on pre-trained static embeddings, our RAG-based framework dynamically retrieves contextually relevant emotional data and generates an Emotion Prompt. This ensures a closer alignment between depression-related text and emotional references, enhancing both accuracy and adaptability.

These findings validate our initial motivation: a dynamic and context-aware retrieval mechanism is more effective than static pre-trained models. By leveraging the RAG framework and the Emotion Prompt, our model effectively bridges the semantic gap between depression and sentiment datasets, contributing to more accurate and interpretable depression severity predictions.

\subsection{Ablation Study}

\begin{table}[h!]
\centering
\caption{Ablation study on the Emotion Prompt.}
\label{tab:ablation_study}
\begin{tabular}{c|cc}
\hline
\textbf{Emotion Prompt} & \textbf{CCC ↑} & \textbf{MAE ↓} \\ 
\hline
           & 0.499 & 4.88 \\
\checkmark & \textbf{0.593} & \textbf{3.95} \\
\hline
\end{tabular}
\end{table}

As shown in Table~\ref{tab:ablation_study}, integrating the Emotion Prompt significantly enhances the model's performance in depression detection. Without the Emotion Prompt, the model achieves a CCC of 0.499 and an MAE of 4.88. When the Emotion Prompt is included, the CCC improves to \textbf{0.593} and the MAE reduces to \textbf{3.95}, demonstrating substantial gains in both evaluation metrics.

The results highlight the key role of the Emotion Prompt in addressing the semantic gap between depression-related text and external sentiment cues. By dynamically retrieving and incorporating contextually relevant emotional information, the Emotion Prompt enables the model to better capture nuanced emotional dependencies and patterns. Unlike static pre-trained embeddings or traditional transfer learning, our approach allows real-time adaptation to task-specific emotional content. This ensures that retrieved emotional cues are contextually aligned with the depression detection task, resulting in enhanced multimodal representation and improved model robustness. These findings validate the effectiveness of our RAG strategy and emphasize the Emotion Prompt's contribution to bridging semantic gaps and enriching emotional understanding, ultimately leading to more accurate and interpretable depression severity predictions.

\section{CONCLUSION}

In this study, we introduced a novel Retrieval-Augmented Generation (RAG) framework for multimodal depression detection, addressing key limitations of existing pre-training and multi-task learning approaches. By dynamically retrieving semantically relevant emotional text and sentiment labels from an external dataset and generating an Emotion Prompt via a Large Language Model (LLM), our method effectively enhances the representation of emotional cues across text, audio, and video modalities. This dynamic integration reduces reliance on static pre-trained embeddings and mitigates domain discrepancies between sentiment and depression datasets. Experimental results on the AVEC 2019 benchmark demonstrate that our approach outperforms state-of-the-art methods in both accuracy and robustness. Moreover, the interpretability provided by the retrieval and prompt generation process offers valuable insights for clinical diagnosis. Future work will explore further optimization of retrieval strategies and broader applications in related affective computing tasks.

\section*{ETHICS STATEMENT}

This study uses the publicly available AVEC 2019 dataset~\cite{ringeval2019avec}, which was collected with informed consent and approved by the corresponding Institutional Review Board. No new data involving human subjects were collected.



\section*{ACKNOWLEDGMENT}

This work is supported in part by JSPS KAKENHI Grant Numbers JP23K16909, JST BOOST, Grant Number JPMJBS2428.

\bibliographystyle{IEEEtran}
\bibliography{references}

\begin{thebibliography}{99}

\bibitem{santomauro2021global}
Santomauro, Damian F., Herrera, Ana M. Mantilla, Shadid, Jamileh, Zheng, Peng, Ashbaugh, Charlie, Pigott, David M., Abbafati, Cristiana, Adolph, Christopher, Amlag, Joanne O., Aravkin, Aleksandr Y., et al.
"Global prevalence and burden of depressive and anxiety disorders in 204 countries and territories in 2020 due to the COVID-19 pandemic."
The Lancet, vol. 398, no. 10312, pp. 1700--1712, 2021.

\bibitem{friedrich2017depression}
Friedrich, Mary Jane.
"Depression is the leading cause of disability around the world."
JAMA, vol. 317, no. 15, pp. 1517--1517, 2017.

\bibitem{sun2021multi}
Sun, Hao, Liu, Jiaqing, Chai, Shurong, Qiu, Zhaolin, Lin, Lanfen, Huang, Xinyin, and Chen, Yenwei.
"Multi-modal adaptive fusion transformer network for the estimation of depression level."
Sensors, vol. 21, no. 14, pp. 4764, 2021.

\bibitem{liu2022computer}
Liu, Jiaqing, Huang, Yue, Chai, Shurong, Sun, Hao, Huang, Xinyin, Lin, Lanfen, and Chen, Yen-Wei.
"Computer-aided detection of depressive severity using multimodal behavioral data."
In *Handbook of Artificial Intelligence in Healthcare: Vol. 1-Advances and Applications*, pp. 353--371, Springer, 2022.

\bibitem{teng2022transformer}
Teng, Shiyu, Chai, Shurong, Liu, Jiaqing, Tateyama, Tomoko, Huang, Xinyin, and Chen, Yen-Wei.
"A transformer-based multimodal network for audiovisual depression prediction."
In *2022 IEEE 11th Global Conference on Consumer Electronics (GCCE)*, pp. 761--764, IEEE, 2022.

\bibitem{sun2022tensorformer}
Sun, Hao, Chen, Yen-Wei, and Lin, Lanfen.
"Tensorformer: A tensor-based multimodal transformer for multimodal sentiment analysis and depression detection."
IEEE Transactions on Affective Computing, vol. 14, no. 4, pp. 2776--2786, 2022.

\bibitem{10444213}
Teng, Shiyu, Chai, Shurong, Liu, Jiaqing, Tateyama, Tomoko, Lin, Lanfen, and Chen, Yen-Wei.
"Multi-Modal and Multi-Task Depression Detection with Sentiment Assistance."
In *2024 IEEE International Conference on Consumer Electronics (ICCE)*, pp. 1--5, IEEE, 2024.

\bibitem{10782904}
Teng, Shiyu, Chai, Shurong, Liu, Jiaqing, Tateyama, Tomoko, Lin, Lanfen, and Chen, Yen-Wei.
"A Sentiment Pre-trained Text-Guided Multimodal Cross-Attention Transformer for Improved Depression Detection."
In *46th Annual International Conference of the IEEE Engineering in Medicine and Biology Society (EMBC)*, pp. 1--4, IEEE, 2024.

\bibitem{zadeh2018multimodal}
Zadeh, AmirAli Bagher, Liang, Paul Pu, Poria, Soujanya, Cambria, Erik, and Morency, Louis-Philippe.
"Multimodal language analysis in the wild: Cmu-mosei dataset and interpretable dynamic fusion graph."
In *Proceedings of the 56th Annual Meeting of the Association for Computational Linguistics (Volume 1: Long Papers)*, pp. 2236--2246, 2018.

\bibitem{lewis2020retrieval}
Lewis, Patrick, Perez, Ethan, Piktus, Aleksandra, Petroni, Fabio, Karpukhin, Vladimir, Goyal, Naman, Küttler, Heinrich, Lewis, Mike, Yih, Wen-tau, and Rocktäschel, Tim, et al.
"Retrieval-augmented generation for knowledge-intensive nlp tasks."
Advances in Neural Information Processing Systems, vol. 33, pp. 9459--9474, 2020.

\bibitem{izacard-grave-2021-leveraging}
Izacard, Gautier, and Grave, Edouard.
"Leveraging Passage Retrieval with Generative Models for Open Domain Question Answering."
In *Proceedings of the 16th Conference of the European Chapter of the Association for Computational Linguistics (EACL)*, pp. 874--880, 2021.

\bibitem{valstar2016avec}
Valstar, Michel, Gratch, Jonathan, Schuller, Björn, Ringeval, Fabien, Lalanne, Denis, Torres Torres, Mercedes, Scherer, Stefan, Stratou, Giota, Cowie, Roddy, and Pantic, Maja.
"Avec 2016: Depression, mood, and emotion recognition workshop and challenge."
In *Proceedings of the 6th International Workshop on Audio/Visual Emotion Challenge*, pp. 3--10, 2016.

\bibitem{ringeval2019avec}
Ringeval, Fabien, Schuller, Björn, Valstar, Michel, Cummins, Nicholas, Cowie, Roddy, Tavabi, Leili, Schmitt, Maximilian, Alisamir, Sina, Amiriparian, Shahin, Messner, Eva-Maria, et al.
"AVEC 2019 workshop and challenge: State-of-mind, detecting depression with AI, and cross-cultural affect recognition."
In *Proceedings of the 9th International on Audio/Visual Emotion Challenge and Workshop*, pp. 3--12, 2019.

\bibitem{openai2023gpt4}
OpenAI.
"GPT-4 Technical Report."
arXiv preprint arXiv:2303.08774, 2023.

\bibitem{johnson2019billion}
Johnson, Jeff, Douze, Matthijs, and Jégou, Hervé.
"Billion-scale similarity search with GPUs."
IEEE Transactions on Big Data, vol. 7, no. 3, pp. 535--547, 2019.

\bibitem{devlin-etal-2019-bert}
Devlin, Jacob, Chang, Ming-Wei, Lee, Kenton, and Toutanova, Kristina.
"BERT: Pre-training of Deep Bidirectional Transformers for Language Understanding."
In *Proceedings of the 2019 Conference of the North American Chapter of the Association for Computational Linguistics (NAACL)*, pp. 4171--4186, 2019.

\bibitem{lawrence1989concordance}
Lawrence, I., and Lin, Kuei.
"A concordance correlation coefficient to evaluate reproducibility."
Biometrics, pp. 255--268, 1989.

\bibitem{kaya2019predicting}
Kaya, Heysem, Fedotov, Dmitrii, Dresvyanskiy, Denis, Doyran, Metehan, Mamontov, Danila, Markitantov, Maxim, Akdag Salah, Alkim Almila, Kavcar, Evrim, Karpov, Alexey, and Salah, Albert Ali.
"Predicting depression and emotions in the cross-roads of cultures, para-linguistics, and non-linguistics."
In *Proceedings of the 9th International on Audio/Visual Emotion Challenge and Workshop*, pp. 27--35, 2019.

\bibitem{rodrigues2019multimodal}
Rodrigues Makiuchi, Mariana, Warnita, Tifani, Uto, Kuniaki, and Shinoda, Koichi.
"Multimodal fusion of bert-cnn and gated cnn representations for depression detection."
In *Proceedings of the 9th International on Audio/Visual Emotion Challenge and Workshop*, pp. 55--63, 2019.

\bibitem{yin2019multi}
Yin, Shi, Liang, Cong, Ding, Heyan, and Wang, Shangfei.
"A multi-modal hierarchical recurrent neural network for depression detection."
In *Proceedings of the 9th International on Audio/Visual Emotion Challenge and Workshop*, pp. 65--71, 2019.

\end{thebibliography}

\end{document}